\documentclass[10pt,twocolumn,letterpaper]{article}

\usepackage{wacv}
\usepackage{times}
\usepackage{epsfig}
\usepackage{graphicx}
\usepackage{amsmath}
\usepackage{amssymb}
\usepackage{booktabs}
\usepackage{svg}
\usepackage{multirow}
\usepackage{comment}
\usepackage{xcolor}
\usepackage{float}
\usepackage[accsupp]{axessibility}

\makeatletter
\def\blfootnote{\xdef\@thefnmark{}\@footnotetext}
\makeatother

\def\wacvPaperID{1781} 

\wacvapplicationstrack 

\wacvfinalcopy 

\ifwacvfinal
\usepackage[breaklinks=true,bookmarks=false]{hyperref}
\else
\usepackage[pagebackref=true,breaklinks=true,colorlinks,bookmarks=false]{hyperref}
\fi

\begin{document}
\title{Improving Deep Facial Phenotyping for Ultra-rare Disorder Verification Using Model Ensembles}

\author{Alexander Hustinx$^1$, 
Fabio Hellmann$^2$, 
Ömer Sümer$^2$, 
Behnam Javanmardi$^1$,\\
Elisabeth André$^2$, 
Peter Krawitz$^1$, 
Tzung-Chien Hsieh$^{1*}$\\
\normalsize $^1$ Institute for Genomic Statistics and Bioinformatics, University Hospital Bonn, University of Bonn\\
\normalsize $^2$ Chair for Human-Centered Artificial Intelligence, University of Augsburg\\
{\tt\small \{ahustinx,b-jav,pkrawitz,thsieh\}@uni-bonn.de}\\
{\tt\small \{fabio.hellmann,oemer.suemer,andre\}@informatik.uni-augsburg.de}
}

\maketitle
\thispagestyle{empty}

\begin{abstract}
Rare genetic disorders affect more than 6\% of the global population. Reaching a diagnosis is challenging because rare disorders are very diverse. Many disorders have recognizable facial features that are hints for clinicians to diagnose patients. Previous work, such as GestaltMatcher, utilized representation vectors produced by a DCNN similar to AlexNet to match patients in high-dimensional feature space to support ``unseen" ultra-rare disorders. However, the architecture and dataset used for transfer learning in GestaltMatcher have become outdated. Moreover, a way to train the model for generating better representation vectors for unseen ultra-rare disorders has not yet been studied. Because of the overall scarcity of patients with ultra-rare disorders, it is infeasible to directly train a model on them. Therefore, we first analyzed the influence of replacing GestaltMatcher DCNN with a state-of-the-art face recognition approach, iResNet with ArcFace. Additionally, we experimented with different face recognition datasets for transfer learning. Furthermore, we proposed test-time augmentation, and model ensembles that mix general face verification models and models specific for verifying disorders to improve the disorder verification accuracy of unseen ultra-rare disorders. Our proposed ensemble model achieves state-of-the-art performance on both seen and unseen disorders. Code is available at \tt\small\href{https://www.github.com/igsb/GestaltMatcher-Arc}{\textcolor{blue}{github.com/igsb/GestaltMatcher-Arc}}.\blfootnote{* Corresponding author.}

\end{abstract}

\section{Introduction}
More than 6\% of the global population is affected by rare genetic disorders~\cite{Ferreira2019-vh}. Because of the rarity and diversity of genetic disorders, reaching a diagnosis is challenging and time-consuming.
More than a third of patients wait for over five years to receive a diagnosis, often referred to as the ``diagnostic odyssey"~\cite{Zurynski2017-kh}. 
Many disorders have distinctive dysmorphic facial features, and these features (gestalt) are hints for clinicians to diagnose patients.
However, recognizing the facial gestalt presented on a patient’s face highly relies on the clinician’s experience, and it is very difficult if the clinician has never seen the disorder before.

\begin{figure}[t]
\begin{center}
   \includegraphics[width=1\linewidth]{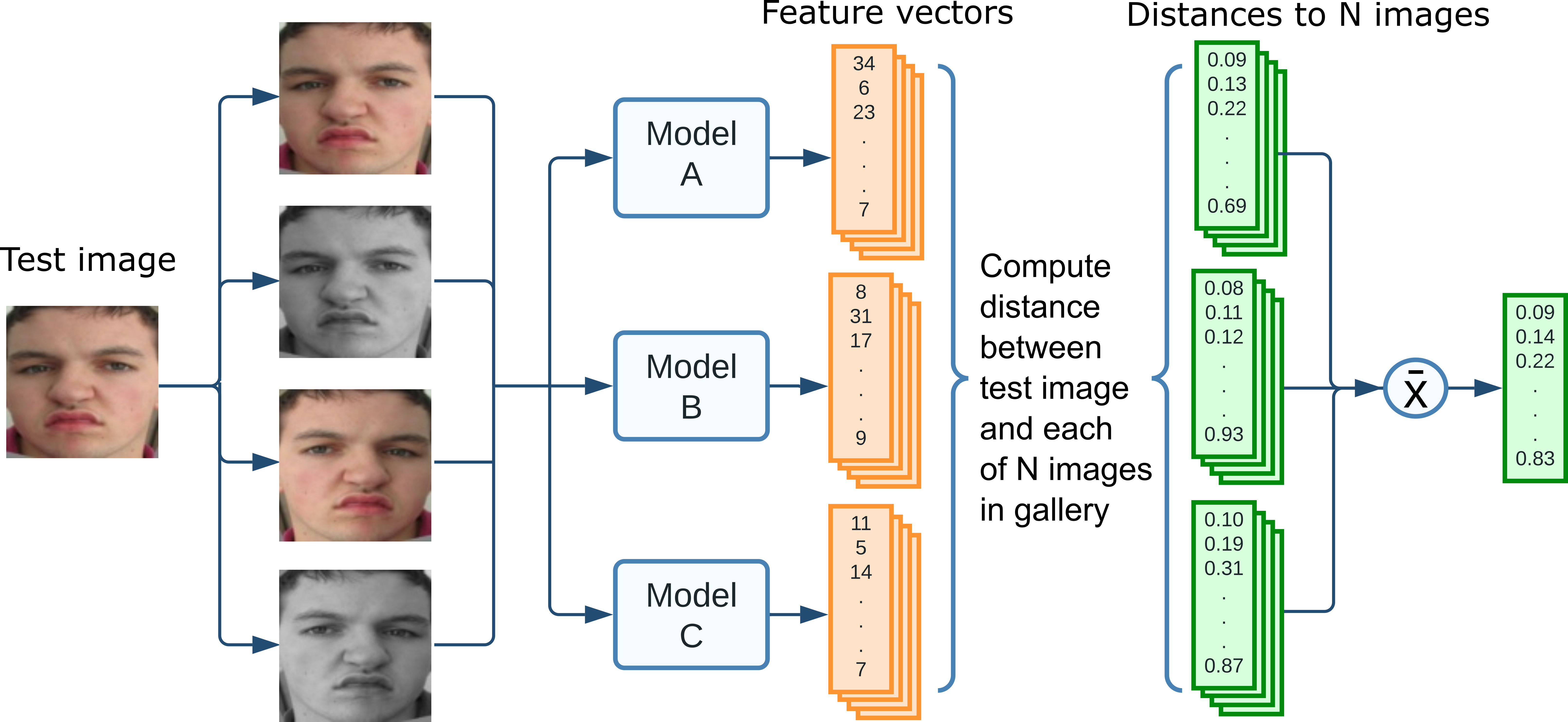}
\end{center}
   \caption{Model ensemble of our approach. We first performed test time augmentation to augment the test image into four images (color and horizontal flip). The four augmented images were further encoded by three different models into 12 representation vectors. We then compared the cosine distance of the 12 representation vectors to the 12 representation vectors from each of the $N$ images in the gallery. It resulted in 12 distance vectors, and each vector contains $N$ cosine distances. In the end, we averaged over 12 distance vectors ($\bar{X}$) to obtain the final distance vector, which further ranked the $N$ images in the gallery. The gallery image with a smaller distance is more similar to the test image.}
\label{fig:workflow}
\end{figure}

With recent advances in computer vision, many next-generation phenotyping (NGP) approaches have emerged to predict rare disorders by analyzing patient's 2D frontal image~\cite{Cerrolaza2016-ln,Dudding-Byth2017-fq,Duong2022-uq,Ferry2014-mm,Gurovich2019-vf,Hong2021-nr,Hsieh2022-ec,Kuru2014-ia,Liehr2018-am,Liu2021-xo,Shukla2017-bz,Van_der_Donk2019-px,Wang2016-re}. Among them, DeepGestalt~\cite{Gurovich2019-vf} utilized transfer learning to train a deep convolutional neural network on CASIA~\cite{Yi2014-wu} and to further fine-tune on over 17,106 patient frontal images with 216 disorders. It achieved 91\% of top-10 accuracy on a test set of 502 images with 92 different disorders and even outperformed human experts. Although DeepGestalt demonstrated extraordinary accuracy in predicting these disorders, it can only classify the disorders it has seen during training, and the trained syndromes are only a tiny proportion of all genetic disorders. If the disorders are ultra-rare or novel, we cannot include them in the model training due to a lack of images. These ``unseen" syndromes are often cases in the real world (Supplementary Figure S1). Therefore, a way to support unseen syndromes becomes crucial.

To support unseen syndromes, GestaltMatcher was proposed as an extension of DeepGestalt that takes the feature layer before the classification layer in DeepGestalt as the encoder that learned facial dysmorphic features~\cite{Hsieh2022-ec}. It encoded the frontal image into a 320-dimensional representation vector. These representation vectors further spanned a feature space. All patients with genetic disorders can be matched or clustered in this space, no longer being limited to the disorders that are trained on (seen) by the networks.

However, both DeepGestalt and GestaltMatcher used the architecture and dataset for transfer learning proposed by Yi \etal~\cite{Yi2014-wu} in 2014. Since then, many larger face recognition datasets~\cite{An2021-iq,Cao2017-ag,Deng2019-bl} and more advanced architectures and loss functions~\cite{Deng2019-bl,noauthor_undated-ev,He2016-jy,Liu2017-mp,Wang2018-tw} were proposed that achieved higher performance on the face verification task. Therefore, the first aim of this study was to update the architecture by using iResNet~\cite{noauthor_undated-ev} and ArcFace~\cite{Deng2019-bl}, and investigating the influence of using different face datasets for transfer learning.

Moreover, a way to train the model that generates better feature representations for unseen ultra-rare disorders has not yet been studied. Hence, the second aim was to investigate different training settings to understand how we can obtain better feature representations for unseen disorders. Our findings showed that fine-tuning on the disorder dataset improved the seen disorder's accuracy but was not always beneficial for the unseen disorders. Thus, we proposed a model ensemble to integrate face verification and disorder models to improve performance on both seen and unseen syndromes (Figure \ref{fig:workflow}).

In summary, the contributions in this paper are as follows:
\vspace*{-0.75em}
\begin{itemize}
\itemsep0em
    \item We analyzed the influence of updating the architecture, loss function, and face dataset used for transfer learning.
    \item We investigated the training settings to generate better feature representations for unseen ultra-rare disorders.
    \item Every updated individual model outperformed the GestaltMatcher baseline model by \cite{Hsieh2022-ec}.
    \item We proposed a model ensemble to mix general face verification models and models specific for verifying disorders to improve the disorder verification accuracy of unseen ultra-rare disorders.
\end{itemize}

All experiments were conducted on the GestaltMatcher Database (GMDB), which is available to medical-related research communities.

\section{Related works}
\subsection{Next-generation phenotyping}
Many rare genetic disorders present recognizable facial features, also called ``facial gestalt". For example, patients with Down syndrome have a distinct facial gestalt. Recognizing the facial gestalt shown in a patient's face is helpful for clinicians in diagnosing the patient. However, it highly relies on the clinician's experience. When disorders are ultra-rare or novel, a clinician has very likely not seen the disorders before. Therefore, next-generation phenotyping approaches that analyze patients 2D frontal face photo to facilitate the diagnosis become crucial.

In 2014, Ferry \etal utilized the shape and appearance representation vectors derived from trained Active Appearance Models and further constructed a feature space they dubbed the ``Clinical Face Phenotype Space" (CFPS) using the representation vectors for disorder classification ~\cite{Ferry2014-mm}. They trained the model on 1,363 images of eight syndromes and 1,515 images from healthy individuals, and it was the first study that analyzed a relatively large cohort.

With the rapid development of computer vision, many approaches using deep convolutional neural networks (DCNN) have been proposed.
Shukla \etal ~\cite{Shukla2017-bz} trained AlexNet~\cite{Krizhevsky_undated-mk} on the entire face and four different facial regions (top right, top left, bottom right, and bottom left) of LFW~\cite{Huang2007-yh} and concatenated five representation vectors into one 20,480 dimensional vector.
In the end, a support vector machine was used to classify six different disorders. Later in 2019, DeepGestalt~\cite{Gurovich2019-vf}, which utilized transfer learning to train a DCNN on more than 17,106 patient photos with 216 different disorders, showed a high prediction accuracy that outperformed clinical experts. Hong \etal ~\cite{Hong2021-nr} also used transfer learning to fine-tune VGG-16~\cite{Simonyan2014-qk} on 228 children with genetic disorders and 228 healthy children. It performed binary classification (with/without a genetic disorder) that could be used for screening.

However, the prevalence of rare disorders is highly imbalanced (Supplementary Figure S1). The number of disorders with enough photos to be included for training the DCNN are a relatively small proportion of all genetic disorders. Syndromes with very few images or novel disorder were not suitable to the classification methods. Therefore, Marbach \etal ~\cite{Marbach2019-su} demonstrated matching two unrelated patients with a novel disease by using facial embeddings encoded by FaceNet~\cite{Schroff2015-jo}. In addition, van der Donk \etal ~\cite{Van_der_Donk2019-px} concatenated the facial embeddings encoded by a normal face recognition model and model trained by disorders. They further performed a clustering analysis to validate the given cohorts with significant facial gestalt. Therefore, generating facial embeddings that generalize dysmorphic facial features for unseen ultra-rare disorders is essential for rare disorder analysis.

\subsection{DeepGestalt}
DeepGestalt was proposed by FDNA Inc., which is considered as the current state-of-the-art disorder classification framework~\cite{Gurovich2019-vf}. It uses the architecture proposed by Yi \etal ~\cite{Yi2014-wu} trained on CASIA~\cite{Yi2014-wu} to learn general facial features as a base for transfer learning, to later fine-tune the network on 17,106 patient images with 216 different disorders.
The architecture, similar to AlexNet, consists of ten convolutional layers, where every two convolutional layers are followed by a pooling layer, and optimizes a Softmax loss function. 

Gurovich \etal proposed an ensemble method that first cropped the face into multiple regions. The aforementioned architecture was used to train a model for each of the facial regions. In the end, it aggregated the softmax values obtained from each region to perform the diagnosis. It showed 91\% of the top-10 accuracy on a test set of 502 images with 92 disorders. In addition to predicting the disorder, it also demonstrated the ability to classify the subtypes of a disorder.

DeepGestalt is used by thousands of clinicians in their daily diagnosis, and is further integrated into the exome sequencing analysis that facilitates the diagnosis on the molecular level~\cite{Hsieh2019-tz}. However, as briefly discussed in the previous section, DeepGestalt does not work on ultra-rare or novel disorders unseen during training. Therefore, GestaltMatcher was proposed to overcome this limitation.

\subsection{GestaltMatcher}
GestaltMatcher~\cite{Hsieh2022-ec} is an extension of the DeepGestalt approach. It used the same architecture and face dataset (CASIA) as a base for transfer learning. After training, it used the last 320-dimensional fully-connected layer before the classification layer as the feature layer, and used it as an encoder that encoded each image into a 320-dimensional representation vector. The representation vectors further spanned a CFPS. In the CFPS, patients with rare disorders can be matched to other similar patients. Moreover, clustering analysis can be performed to analyze the similarity among different disorders. GestaltMatcher has been used in several studies to analyze patient similarities~\cite{Asif2022-iv,Ebstein2021-rk,Guo2022-os}.

The advantage of GestaltMatcher is that it is no longer limited to the disorders it has seen during training, and it enables researchers to quantify patient-to-patient or syndrome-to-syndrome similarity. However, GestaltMatcher used the same architecture and pre-trained dataset as DeepGestalt that are relatively outdated. Therefore, a study is required to update the architecture and explore methods that improve the performance of unseen ultra-rare disorders verification.

\begin{table}
  \begin{center}
    {\small{
\begin{tabular}{lrr}
\toprule
Dataset & \# of images & \# of individuals \\
\midrule
VGG2~\cite{Cao2017-ag} & 3.31M &  9,131 \\
CASIA~\cite{Yi2014-wu} & 0.49M & 10,575 \\
MS1MV2~\cite{Deng2019-bl} & 5.8M & 85K \\
MS1MV3~\cite{Deng2019-bl} & 5.1M & 93K \\
Glint360K~\cite{An2021-iq} & 17M & 360K\\
\bottomrule
\end{tabular}
}}
\end{center}

\caption{Overview of the face datasets.}
\label{tab:overview}
\end{table}

\section{Datasets and methodology}
\subsection{Datasets}
\subsubsection{Face recognition datasets}
In this paper, we experimented with five different face recognition datasets to be used for training the (transfer learning) base model: VGG2~\cite{Cao2017-ag}, CASIA~\cite{Yi2014-wu}, MS1MV2~\cite{Deng2019-bl}, MS1MV3~\cite{Deng2019-bl}, and Glint360K~\cite{An2021-iq}. The full name of CASIA dataset is CASIA-WebFace. We used CASIA as abbreviation in this paper. The number of images in the datasets ranges from 0.49M to 17M. An overview of the datasets is shown in Table ~\ref{tab:overview}. 

\subsubsection{GestaltMatcher Database - rare disorder dataset}
Hsieh \etal~\cite{Hsieh2022-ec} built up GestaltMatcher Database\footnote{\url{https://db.gestaltmatcher.org/}} (GMDB), which collects medical images of rare disorder from publications and patients with proper consent from clinics. It is open to clinicians and researchers working in medical research fields. To avoid data abuse, applicants need to be reviewed by a committee of GMDB before they can access the database.

We used GMDB (v1.0.3) to fine-tune the base models on faces of patients with disorders. GMDB (v1.0.3) contains 7,459 frontal images of 5,995 patients with 449 different disorders. All the disorders have at least two patients. The dataset was further divided into two sets, a ``frequent" (GMDB-Frequent) and a ``rare" (GMDB-Rare) set. The disorders with more than six patients were assigned to GMDB-Frequent, while the disorders with six or fewer patients were assigned to GMDB-Rare.

\begin{figure}[t]
\begin{center}
   \includegraphics[width=1\linewidth]{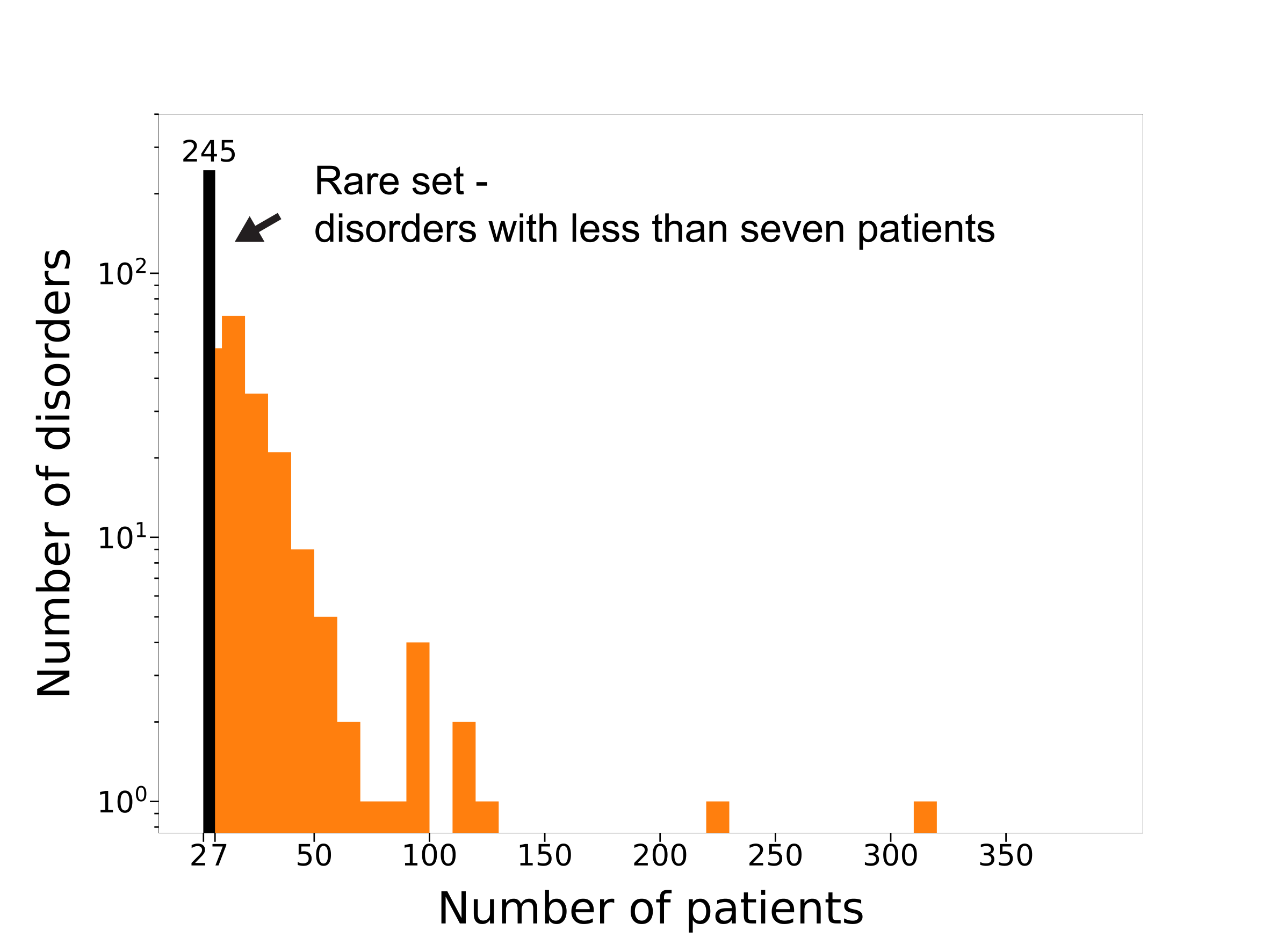}
\end{center}
   \caption{Disorder distribution of GMDB. The X-axis shows the number of patients in the disorder. The Y-axis shows the number of disorders with the corresponding number of patients in X-axis, and it is in log scale. The black bar is the rare set (GMDB-Rare), that has disorders with more than one patient and fewer than seven patients.}
\label{fig:gmdb_distribution}
\end{figure}

There are 6,354 images of 5,123 patients with 204 disorders in GMDB-Frequent and 1,105 images of 872 patients with 245 disorders in GMDB-Rare. We fine-tuned the base models on GMDB-Frequent, thus disorders in this set can be considered as ``seen" disorders. For training, GMDB-Frequent was further divided into 5,100 images for the training set, 661 images for the validation set, and 593 images for the test set. On the other hand, GMDB-Rare was ``unseen" during training. We used GMDB-Rare to simulate ultra-rare or novel disorders in real-world scenarios. The overview of the GMDB dataset is shown in Table ~\ref{tab:GMDB}. In Figure ~\ref{fig:gmdb_distribution}, GMDB shows a long tail distribution. GMDB-Rare has only 14.5\% (872/5995) of all patients, but it covers 54.5\% of the disorders. The distribution of GMDB is similar to the estimation of disorder prevalence in the real world (Supplementary Figure S1).

\begin{table}
  \begin{center}
    {\small{
\begin{tabular}{lrrr}
\toprule
Dataset & \# of images & \# of patients & \# of disorders\\
\midrule
GMDB-Frequent & 6,354 & 5,123 & 204 \\
GMDB-Rare & 1,105 &  872 & 245  \\ 
\midrule
Total & 7,459 &  5,995 & 449 \\
\bottomrule
\end{tabular}
}}
\end{center}

\caption{Overview of GMDB dataset. GMDB-Frequent is used for fine-tuning and thus ``seen'' by the model, while disorders in GMDB-Rare are ``unseen" to the model.}
\label{tab:GMDB}
\end{table}

\subsection{Evaluation}
We evaluated the performance of the base models on the popular face verification dataset Labeled Faces in the Wild (LFW) \cite{Huang2007-yh}. During evaluation, two faces were compared, and the models verified whether they belong to the same person. The evaluation set consisted of 11-folds, where the first was used to establish a threshold, while the remaining 10-folds were used for the final evaluation.

Most importantly, we used GMDB to evaluate the base models, our fine-tuned models, and eventually the model ensemble. During the evaluation procedure, the feature space was populated with a gallery set of diagnosed patients' feature vectors, being either the seen disorders (GMDB-Frequent), the unseen disorders (GMDB-Rare) or an unified gallery (GMDB-Frequent and -Rare). Afterwards, representation vectors of test images were matched to the gallery cases in the feature space. For the unseen test images, 10-fold cross-validation was performed.

We first calculated the cosine distances between the test image and each image in the gallery. The cosine distance further ranked the gallery images. Hsieh \etal \cite{Hsieh2022-ec} showed the top-$k$ ($k \in [1,5,10,30]$) mean accuracy (as described in Equation \ref{eq:ma}) for test images of seen and unseen disorders. Instead, we focused on the top-1 and top-5 results in this paper, though top-10 and top-30 are included in the Supplemental Tables. The GestaltMatcher DCNN\footnote{\url{https://github.com/igsb/GestaltMatcher}} (GM-Hsieh2022) was used as the baseline, having retrained it on the recent version of GMDB.

\begin{equation}
    mA_k = \frac{1}{C} \sum^C_c A_{k,c} ~,
    \label{eq:ma}
\end{equation}
where $mA_k$ is the top-$k$ mean accuracy, $C$ is the number of classes, $c$ is the class index, and $A_{k,c}$ is the top-$k$ accuracy for class $c$. 

\subsection{Model architecture and training}
Our model architecture is based on the one used by Deng \etal ~\cite{Deng2019-bl}. They used a popular variation on the ResNet architecture named iResNet~\cite{noauthor_undated-ev}. It includes more batch normalization. In addition to the original implementation, it replaces the ReLU activation function with PReLU, and lastly, it batch normalizes the computed representation vector.

The training procedure was split into two steps:
\vspace*{-0.75em}
\begin{itemize}
\itemsep0em
    \item Training a base model on a face recognition dataset for transfer learning;
    \item Fine-tuning that base model on GMDB for disorder verification.
\end{itemize}

For the first part, we used pre-trained models supplied by insightface\footnote{\url{https://github.com/deepinsight/insightface}} that have been trained on different face recognition datasets using Additive Angular Margin Loss (ArcFace). The loss is defined by the Equation \ref{eq:arcface}.
\begin{equation}
    L_{Arc} = -\frac{1}{N}\sum_{i=1}^{N}log\frac{e^{s \cos(\theta_{y_{i}}+m)}}{e^{s \cos(\theta_{y_{i}}+m)} + \sum _{j=1, j\neq y_{i}}^{n}e^{s \cos\theta _{j}}} ~,
    \label{eq:arcface}
\end{equation}

where $\theta_j$ is the angle between weights $W_j$ and feature $x_i$. We insert angular margin $m$ to get a new angle between the true logit, $y_i$, and our representation vector to become $\theta_{y_i} + m$. $s$ is the scale of $L_2$ normalized representation vectors. The pre-trained models used $m$ and $s$ set to 0.5 and 64, respectively. Representations learned using this loss tended to have stronger distinctions between different classes and better similarities for the same classes than most other metric learning losses.

An important preprocessing step of insightface's training procedure is to align faces based on five landmarks: left and right eye, nose, left and right mouth corner. This alignment is essential to reproduce their performance. For our implementation, we used RetinaFace~\cite{Deng2020-nw} to obtain these landmarks and the alignment code they supplied, which uses an affine transformation based on matching landmark locations.

For the second part, we made minor changes to the model architecture. We removed the batch normalization of the computed features. This normalization was necessary for ArcFace, which we did not use during fine-tuning. Instead, due to the small dataset size, significant class imbalance, and long tail distributions, we decided only to optimize Weighted Cross Entropy Softmax Loss (WCE). 

To address the class imbalance, we used Equation \ref{eq:wc} to calculate the class weights, casting them into the range $(0.5, ..., 1.0]$.
\begin{equation}
    W_c = \frac{0.5 \cdot min(D)}{D_c} + 0.5 ~,
    \label{eq:wc}
\end{equation}

where $D$ is the set of frequencies per class, $c$ is the class, and $W_c$ is the WCE weight for class $c$.
If not for our lower bound of $W_c > 0.5$, due to the long-tailed distribution it would be possible for $W_c$ to be lower than $0.01$. This would make the training process challenging.
We also replaced the final fully connected layer to train a classifier on the disorders of the training set. Lastly, we freezed all model weights except those of the feature and classification layer.

We fine-tuned our model on GMDB using aligned faces of size 112x112, randomly flipping horizontally, randomly converting color images to gray, color jittering, and randomly adding zooming/cropping artifacts. We further used the Adam optimizer with a base learning rate of 1e-3, which was reduced by a factor of 2 when the top-5 mean accuracy on the validation set plateaus until convergence. The mean accuracy was calculated with Equation \ref{eq:ma}.
Code is available at \href{https://www.github.com/igsb/GestaltMatcher-Arc}{\texttt{\small\textcolor{blue}{github.com/igsb/GestaltMatcher-Arc}}}.

\subsection{Inference strategy}
\label{sec:inference}
An essential part of our method was our strategy during inference. We aimed to improve our performance on both seen and unseen disorders by computing multiple representation vectors per image, aiming to end up with a better overall ranking than each separate representation vector. We employed two approaches to obtain multiple representation vectors per image: model ensembles and test time augmentation.

\subsubsection{Model ensembles}
Model ensembles are mixtures of models that combine each model’s output. This approach helps achieve a better overall generalization as it leverages each model’s strengths to alleviate the others’ weaknesses. In our case, we presented each model with the same image, computed each model’s representation vector, and averaged the cosine distances from the image to the GMDB gallery set.

For our ensemble, we considered both models that are fine-tuned for disorders and models built for face verification. The face verification models produced strong general features that can be leveraged to verify unseen disorders, while the fine-tuned models were fitted towards features of seen disorders they have been trained on. More specifically, we included one face verification model, one deeper disorder model (using iResNet-100), and one disorder model (using iResNet-50) designed to be less prone to overfitting on the seen disorders. More detailed information on the selected models can be found in Section \ref{sec:exp&res}.

\subsubsection{Test time augementation}
Test time augmentation (TTA), similarly to model ensembles, combines outputs to achieve more robust performance. However, instead of presenting different models with the same image, it presented the same model with an image and augmented versions of that image (e.g., horizontally flipped, converted from color to gray, rotation, and translation).
Ideally, the representation vectors would be close to identical because they are from the same image, and the actual face does not change. In practice, this is usually not the case. This helps average out the cosine distance between the gallery and test set.
 
Of course, not all augmentations make sense to use during TTA. Any augmentation that changes the face structure or affects the required face alignment is potentially harmful for our implemention. Generally, augmentations used during training are well suited for TTA. As such, we used horizontal flipping and conversion from color to gray. 

Finally, we averaged the cosine distance of all models in the ensemble and TTAs per model (i.e., three models and two TTA each, 3x2x2=12 cosine distances). The order of the disorders ranked for verification was determined by the $k$-nearest neighbors of the average cosine distance between the gallery images to the test image. We decided to use $k=1$ due to the highly imbalanced data in GMDB (and thus also in the gallery set). Using a $k > 1$ would be problematic for disorders with only one occurrence in the gallery set. Figure ~\ref{fig:workflow} gives a simplistic view of how these inference strategies work.

\section{Experiments and results}
\label{sec:exp&res}
\subsection{Updating the architecture and base optimization}
We hypothesized that replacing GestaltMatcher's base model (GM-Hsieh2022) with a state-of-the-art face verification model will improve the overall performance on LFW and GMDB.

First, we compared the performance of the GM-Hsieh2022 model, using an AlexNet-like architecture and cross-entropy loss, to iResNet-50 with ArcFace, both trained on CASIA. Afterward, we fine-tuned these models on GMDB. During the fine-tuning process, both models only used weighted cross entropy. Results of the base models and fine-tuned models are shown in Table \ref{tab:comparison_old_new}. An extended version of the table can be found in Supplementary Table S1, and the performance when using the unified gallery in Supplementary Table S5.

We find that the features generated by the ArcFace base model are generally more descriptive than those of the GM-Hsieh2022 base model. This is supported by the higher LFW performance and the overall higher performance on GMDB without fine-tuning. In Table \ref{tab:comparison_old_new}, the LFW accuray is increased from 93.8\% to 98.4\%, and the top-1 and top-5 accuracies for both seen (GMDB-Frequent) and unseen (GMDB-Rare) disorders are improved when we update the model from GM-Hsieh2022 to ArcFace-r50.

After fine-tuning, the GM-Hsieh2022 model improved on both GMDB-Frequent and GMDB-Rare. The ArcFace model significantly increased the performance on seen disorders while decreasing the performance on unseen disorders. We believed this indicated that the model has a higher tendency to overfit on the small dataset than the GM-Hsieh2022 model had. Although the performance of GMDB-Rare dropped after fine-tuning the new model (ArcFace-r50*), the top-1 and top-5 accuracies were still similar to the fine-tuned GestaltMatcher by \cite{Hsieh2022-ec} (GM-Hsieh2022*).

\begin{table}[!htbp]
  \begin{center}
    {\small{
\begin{tabular}{l@{\hskip4pt}c@{\hskip4pt}c@{\hskip4pt}c@{\hskip4pt}c@{\hskip4pt}c@{\hskip4pt}c@{\hskip4pt}}

\toprule
\multirow{2}{*}{Model} & \multirow{2}{*}{LFW} & \multicolumn{2}{c}{GMDB-Frequent} & \multicolumn{2}{c}{GMDB-Rare} \\
& & Top-1&Top-5 & Top-1&Top-5 \\
\midrule
GM-Hsieh2022 & 93.8\% & 10.99\% & 29.39\% & 14.64\% & 27.03\% \\
GM-Hsieh2022* & - & \textbf{15.96\%} & \textbf{33.83\%} & \textbf{19.26\%} & \textbf{36.28\%} \\
ArcFace-r50 & 98.4\% & 21.84\% & 40.87\% & \textbf{22.74\%} & \textbf{37.35\%} \\
ArcFace-r50* & - & \textbf{35.37\%} & \textbf{53.25\%} & 19.29\% & 36.00\% \\
\bottomrule
\end{tabular}
}}

\end{center}
\caption{Comparison of the performance of the GM-Hsieh2022 model and the ArcFace-r50 model on LFW and GMDB. Both have been pre-trained on CASIA and models marked with (*) have been fine-tuned on GMDB. For each column, the best accuracy between the models before fine-tuning and after fine-tuning is boldfaced.}
\label{tab:comparison_old_new}
\end{table}

\begin{figure}[!htbp]
\begin{center}
   \includegraphics[width=1\linewidth]{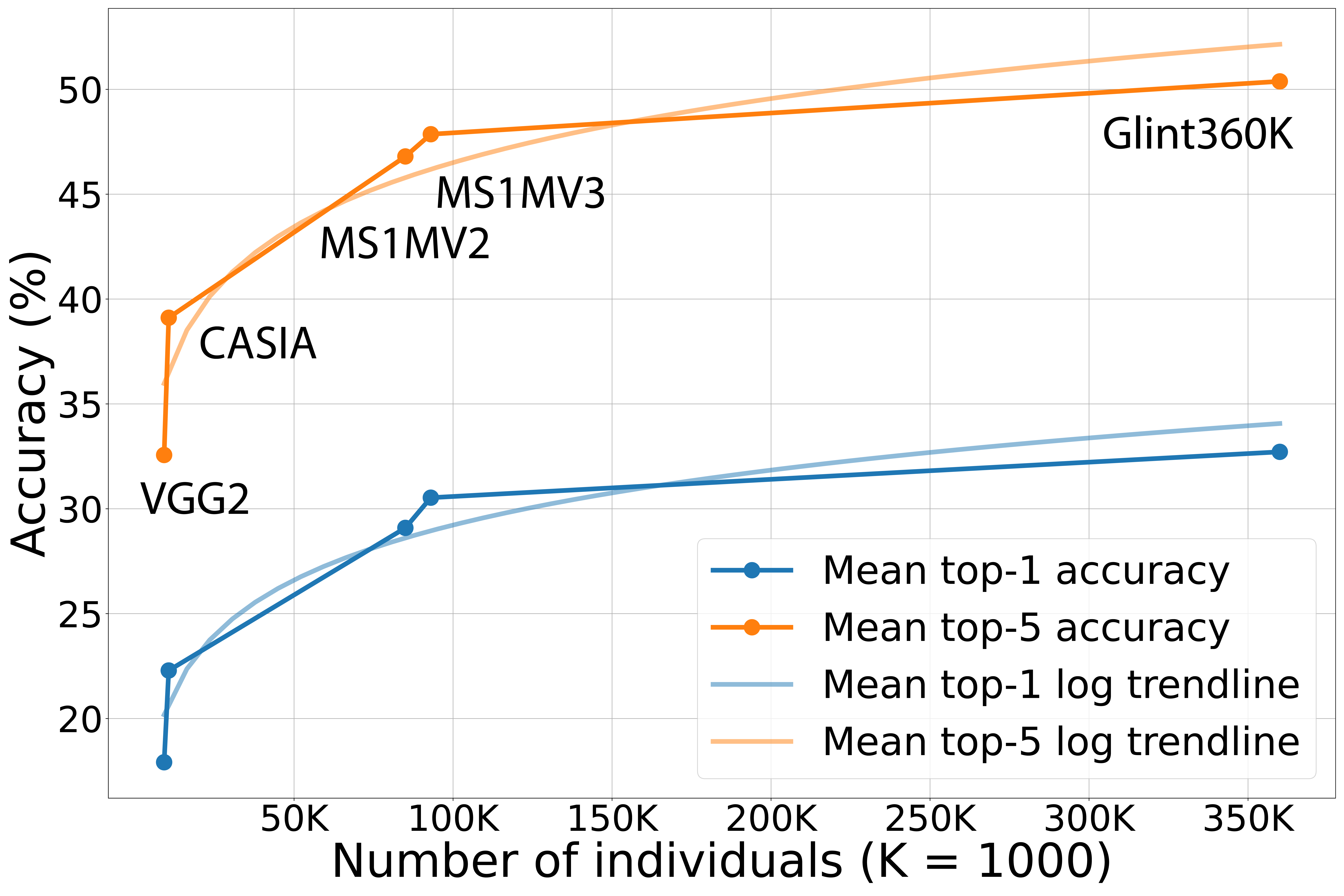}
\end{center}
   \caption{Mean accuracy of the ArcFace-r50 base model on GMDB when using different datasets. The X-axis shows the number of individuals in the datasets. The Y-axis shows the mean accuracy (both GMDB-Frequent and -Rare) of models using different base datasets. The light orange and blue lines shows the logarithmic relation.}
\label{fig:num_individual_acc}
\end{figure}

\begin{table*}
  \begin{center}
    {\small{
\begin{tabular}{l@{\hskip4pt}lllll}
\toprule
\multirow{2}{*}{Dataset} & \multirow{2}{*}{LFW} & \multicolumn{2}{l}{GMDB-Frequent} & \multicolumn{2}{l}{GMDB-Rare} \\
& & Top-1&Top-5 & Top-1&Top-5 \\
\midrule
VGG2 & 98.5\% & 15.52\% & 31.56\% & 20.31\% & 33.57\% \\
CASIA & 98.4\% & 21.84\% & 40.87\% & 22.74\% & 37.35\% \\
MS1MV2 & 99.0\% & 29.14\% & 48.86\% & 29.04\% & 44.74\% \\
MS1MV3 & 98.9\% & 31.54\% & 49.36\% & 29.52\% & 46.36\% \\
Glint360K & 99.0\% & 32.43\% & 53.14\% & 33.00\% & 47.62\% \\
\hline
VGG2* & 85.8\% \textcolor{red}{(-12.7\%)}& 27.50\% \textcolor{blue}{(+11.98\%)} & 49.92\% \textcolor{blue}{(+18.36\%)} & 17.56\% \textcolor{red}{(-2.75\%)}& 33.41\% \textcolor{red}{(-0.16\%)} \\
CASIA* & 75.7\% \textcolor{red}{(-22.7\%)}& 35.37\% \textcolor{blue}{(+13.53\%)} & 53.25\% \textcolor{blue}{(+12.38\%)} & 19.29\% \textcolor{red}{(-3.45\%)} & 36.00\% \textcolor{red}{(-1.35\%)}\\
MS1MV2*& 84.1\% \textcolor{red}{(-17.7\%)}& 39.98\% \textcolor{blue}{(+10.84\%)} & 59.81\% \textcolor{blue}{(+10.95\%)} & 21.86\% \textcolor{red}{(-7.18\%)} & 39.89\% \textcolor{red}{(-4.85\%)} \\
MS1MV3* & 76.4\% \textcolor{red}{(-22.5\%)}& 45.06\% \textcolor{blue}{(+13.52\%)} & 64.64\% \textcolor{blue}{(+15.28\%)} & 24.31\% \textcolor{red}{(-5.21\%)} & 40.28\% \textcolor{red}{(-6.08\%)} \\
Glint360K* & 84.9\% \textcolor{red}{(-12.6\%)}& 41.58\% \textcolor{blue}{(+9.15\%)} & 62.60\% \textcolor{blue}{(+9.46\%)} & 26.55\% \textcolor{red}{(-6.45\%)} & 42.69\% \textcolor{red}{(-4.93\%)} \\
\bottomrule
\end{tabular}
}}

\end{center}
\caption{Comparison of the performance of the ArcFace-r50 models trained on a variety of face recognition datasets. The percentages within parentheses indicate the change between the face verification (base) and the fine-tuned models. For example, the top-1 accuracy on GMDB-Frequent increased by 11.98\% (15.52\% $\rightarrow$ 27.50\%) from VGG2 to VGG2*. Models marked with (*) have been fine-tuned on GMDB.}
\label{tab:comparison_performance_dataset}
\end{table*}

\begin{table}[!htbp]
  \begin{center}
    {\small{
\begin{tabular}{l@{\hskip4pt}c@{\hskip4pt}c@{\hskip4pt}c@{\hskip4pt}c@{\hskip4pt}c@{\hskip4pt}c}
\toprule
\multirow{2}{*}{Model} & \multirow{2}{*}{LFW} & \multicolumn{2}{c}{GMDB-Frequent} & \multicolumn{2}{c}{GMDB-Rare} \\
& & Top-1&Top-5 & Top-1&Top-5 \\
\midrule
r50 & 84.9\% & 41.58\% & 62.60\% & 26.55\% & 42.69\% \\
r50-D/O & 86.2\% & \textbf{46.95\%} & \textbf{66.07\%} & 28.85\% & 45.36\% \\
r50-D/O\textdagger &\textbf{87.6} \% & 44.33\% & 65.76\% & \textbf{29.06\%} & \textbf{46.35\%} \\
\hline
r100 & 91.0\% & 47.96\% & 68.87\% & 26.03\% & 42.22\% \\
r100-D/O & 91.1 \% & 48.37\% & \textbf{71.78\%} & 28.02\% & 44.32\% \\
r100-D/O\textdagger & \textbf{93.0\%} & \textbf{49.25\%} & 69.95\% & \textbf{30.33\%} & \textbf{47.85\%} \\
\bottomrule
\end{tabular}
}}

\end{center}
\caption{Comparison of the performance of iResNet-50 and -100 fine-tuned on GMDB. D/O indicates an additional dropout layer and (\textdagger) indicates the use of $L_2$ weight decay on the feature layer. For each column, the best accuracy among the models (without regularizaton, D/O, and D/O\textdagger) is boldfaced.}
\label{tab:regularlization}
\end{table}

\begin{table*}[!htbp]
  \begin{center}
    {\small{
\begin{tabular}{lccccccc}
\toprule
\multirow{2}{*}{Model} & \multirow{2}{*}{Dataset} & \multirow{2}{*}{Loss} & \multicolumn{2}{c}{GMDB-Frequent} & \multicolumn{2}{c}{GMDB-Rare} \\
& & & Top-1&Top-5 & Top-1&Top-5 \\
\midrule
GM-Hsieh2022 & CASIA* &  CE  & 15.96\% & 33.83\% & 19.26\% & 36.28\% \\
\midrule

r50-D/O\textdagger & Glint360K* &  CE & 44.33\% & 65.76\% & 29.06\% & 46.35\% \\
r50-D/O\textdagger + TTA & Glint360K* & CE & 47.73\% & 67.67\% & 30.29\% & 46.38\% \\
r100-D/O & Glint360K* & CE & 48.37\% & \textbf{71.78\%} & 28.02\% & 44.32\% \\
r100-D/O + TTA & Glint360K* & CE & 51.16\% & 69.58\% & 27.92\% & 46.26\% \\
r100 & Glint360K &  ArcFace & 30.25\% & 54.81\% & 33.25\% & 50.22\% \\
r100 + TTA& Glint360K & ArcFace & 35.25\% & 56.52\% & 33.47\% & 51.61\% \\
\midrule
Model ensemble & n/a & n/a & 52.06\% & 70.70\% & 34.93\% & 52.78\% \\
Model ensemble + TTA & n/a  & n/a & \textbf{52.99\%} & 71.01\% & \textbf{35.98\%} & \textbf{53.93\%} \\
\bottomrule
\end{tabular}
}}

\end{center}
\caption{Comparison of the performance of the GM-Hsieh2022 (baseline) model, two ArcFace models fine-tuned on GMDB, one ArcFace face verification model, and our model ensemble using the three ArcFace models. TTA indicates the model was evaluated using test time augmentation, (*) indicates the model was fine-tuned on GMDB, (D/O) indicates an additional dropout layer, and (\textdagger) indicates the use of $L_2$ weight decay on the feature layer.}
\label{tab:ensemble}
\end{table*}

\subsection{Updating the transfer learning dataset}

We hypothesized that increasing the number of individuals in the transfer learning base dataset will result in better/more general (facial) feature descriptors. However, we expected some drop-off in the performance gain when increasing the number of individuals indefinitely.
To test this hypothesis, we compared the performance on LFW and GMDB with five well-known face recognition datasets: VGG2, CASIA, MS1MV2, MS1MV3, and Glint360K. The results are shown in Table \ref{tab:comparison_performance_dataset} and Figure \ref{fig:num_individual_acc}. An extended version of the table can be found in Supplementary Table S2, and the performance when using the unified gallery in Supplementary Table S6.

Figure \ref{fig:num_individual_acc} shows the average accuracy per base dataset concerning the number of unique individuals in the dataset. Table \ref{tab:comparison_performance_dataset} shows that the ArcFace-r50 base models trained on datasets with more different individuals tends to achieve higher accuracy on both GMDB-Frequent and GMDB-Rare before fine-tuning. We also found a drop-off in accuracy gained when increasing the number of individuals based on the logarithmic relation shown in the Figure \ref{fig:num_individual_acc}.

Moreover, we found that the accuracy on GMDB-Frequent after fine-tuning did not always improve when using a larger dataset. For example, in Table \ref{tab:comparison_performance_dataset}, the top-1 and top-5 accuracies of Glint360K* are lower than the accuracies of MS1MV3*, which drop from 45.06\% to 41.58\% and 64.64\% to 62.60\%, respectively. However, the accuracy on GMDB-Rare after fine-tuning always improved when we used a larger dataset for training the ArcFace-r50 base model.

The results show that both GMDB-Frequent and GMDB-Rare benefit from using a larger dataset for training the ArcFace-r50 base model, especially for the unseen disorders (GMDB-Rare). In addition, it might not be necessary to use a face recognition dataset larger than Glint360K, as the performance gain seems to saturate when the number of individuals in the dataset is larger than 1M.

\subsection{Influence of fine-tuning ArcFace on GMDB}
In an earlier experiment, we saw that fine-tuning ArcFace on GMDB reduced the accuracy on unseen disorders (GMDB-Rare). We believed that fine-tuning the feature representation layer on GMDB will negatively influence the general feature descriptors’ quality by (over)fitting on the small imbalanced dataset, not just when using CASIA as the base dataset but also for larger base datasets. We believed this should reflect in the accuracy on LFW.
As such, we fine-tuned the ArcFace models trained on the CASIA, VGG2, MS1MV2, MS1MV3, and Glint360K on GMDB, and afterward evaluated them on LFW and GMDB. The results are shown in Table \ref{tab:comparison_performance_dataset}. An extended version of the table can be found in Supplementary Table S2, and the performance when using the unified gallery in Supplementary Table S6.

Based on the results in the Table \ref{tab:comparison_performance_dataset}, we find that fine-tuning decreases the performance on unseen disorders (GMDB-Rare) for every model, as well as the general face verification performance on LFW. However, the performance of the models using a larger base dataset still outperform the baseline for these unseen disorders.

\subsection{Additional regularization during fine-tuning to improve generalizability on unseen disorders}
\label{regularization}
We believed that the overfitting, shown in the previous experiment as a drop in accuracy for unseen disorders, can be reduced by adding additional regularization to the feature layer in the form of $L_2$ weight decay and dropout.
We fine-tuned the iResNet-50 and iResNet-100 ArcFace models pre-trained on Glint360K to include additional dropout and additional $L_2$ weight decay of the feature layer ($\lambda=5e^{-5}$). The results are shown in Table \ref{tab:regularlization}. An extended version of the table can be found in Supplementary Table S3, and the performance when using the unified gallery in Supplementary Table S7.

We find that the use of additional dropout improves accuracy for both the seen and unseen disorders. Additional $L_2$ weight decay on the feature layer helps to maintain some of the LFW performance and, on some occasions, is an improvement to dropout. For example, the top-1 accuracy on GMDB-Frequent increases from 48.37\% to 49.25\% when applying $L_2$ weight decay to r100-D/O. Although the improvement of $L_2$ weight decay on seen disorders (GMDB-Frequent) is inconclusive, the improvement on the unseen ones (GMDB-Rare) is clear.

\subsection{Influence of inference strategies}
We believed the inference strategies discussed in Section \ref{sec:inference} will improve most models' accuracy. We hypothesized that presenting our model with an image and slight variations of that image will increase the robustness of the clustering. On top of that, combining our disorder models, fine-tuned on GMDB, with general face verification models will improve generalizability and robustness for both seen and unseen disorders.
Table \ref{tab:ensemble} shows the performance of the baseline (GM-Hsieh2022), each model used in the ensemble with and without TTA, and the model ensemble with and without TTA. An extended version of the table can be found in Supplementary Table S4, and the performance when using the unified gallery in Supplementary Table S8.

In Table \ref{tab:ensemble}, we find that TTA increases almost every test group’s performance. Only the top-5 accuracy on GMDB-Frequent and top-1 accuracy on GMDB-Rare for r100 with dropout (r100-D/O) are decreased after applying TTA. 
Moreover, the model ensemble outperforms every single model in almost every test group, except the top-5 accuracy on GMDB-Frequent. The top-5 accuracy on GMDB-Frequent for the model ensemble is 70.70\% which is slightly lower than 71.78\% from r100 with dropout (r100-D/O). In the end, combining the model ensemble and TTA further improves the performance, achieving state-of-the-art. When comparing the model ensemble with TTA to the GM-Hsieh2022 model, the top-1 accuracy improves from 15.96\% to 52.99\% and 19.26\% to 35.98\% on GMDB-Frequent and GMDB-Rare, respectively, showing a strong performance on both seen and unseen disorders.

\section{Conclusion and future works}
We found that using face recognition datasets with more individuals led to more generalized representation vectors, which in turn form a good base for transfer learning.
Fine-tuning the transfer learning datasets with ArcFace iResNet on GMDB led to a significant increase in performance on seen disorders and a decrease in performance on unseen disorders. The latter is likely caused by overfitting on the seen disorders, which unlearns general facial features.
The use of regularization techniques such as dropout and $L_2$ weight decay can help reduce the impact of overfitting, increasing the performance of unseen disorders.
Moreover, using TTA increased the performance of all models.
Next, combining one face verification model and two disorder verification models in a model ensemble allowed us to leverage their strengths on both seen and unseen disorders.

In conclusion, each model with and without TTA, and our model ensemble outperformed GM-Hsieh2022, where the model ensemble achieves state-of-the-art performance. We believe this work can function as a strong baseline for future comparison in this emerging field.

In this study, we focused on the imbalanced number of patients among the disorders. However, Lumaka \etal reported that the performance of DeepGestalt was biased by the imbalance of ethnicity groups in the training set~\cite{Lumaka2017-ks}. Therefore, an approach that consider the imbalance of ethnicity, sex, and age is important.

Moreover, we only discussed the iResNet with ArcFace and cross entropy. Benchmarking on different architectures and loss functions, such as EfficientNet~\cite{Tan_undated-iq}, CosFace~\cite{Wang2018-tw}, and SphereFace~\cite{Liu2017-mp} is required to understand more on how to obtain more generalized representation vectors for unseen disorders. Besides, using different representation vector dimensions and dimension reduction methods are also a possibility to further optimize the feature representation for unseen disorders.


{\small
\bibliographystyle{ieee_fullname}
\bibliography{paperpile}
}

\end{document}